\documentclass[lettersize,journal]{IEEEtran}
\usepackage{amsmath,amsfonts}
\usepackage{algorithmic}
\usepackage{algorithm}
\usepackage{array}
\usepackage[caption=false,font=normalsize,labelfont=sf,textfont=sf]{subfig}
\usepackage{textcomp}
\usepackage{stfloats}
\usepackage{url}
\usepackage{verbatim}
\usepackage{graphicx}
\usepackage{cite}
\usepackage{makecell}
\begin{document}

\title{
DualTalker: A Cross-Modal Dual Learning Approach for Speech-Driven 3D Facial Animation
%Speech-driven Facial Animation via Cross-modal Dual Learning and Consistency Mapping
% inter-modal dual learning
% cross-modal consistency learning
% Speech-driven facial animation
}

\author{Guinan Su, Yanwu Yang ~\IEEEmembership{ Member,~IEEE}, Zhifeng Li, ~\IEEEmembership{Senior Member,~IEEE}

\thanks{Guinan Su and Zhifeng Li are with the Tencent Data Platform, Shenzhen 518057, China. (E-mail: guinansu33@gmail.com, and michaelzfli@tencent.com}
\thanks{Yanwu Yang is with the School of Electronics and Information Engineering, Harbin Institute of Technology at Shenzhen, Shenzhen, China, and the Peng Cheng Laboratory, Shenzhen, Guangdong, China. (e-mail: 20b952019@stu.hit.edu.cn)}
\thanks{Corresponding author: Zhifeng Li}
}




\maketitle

\begin{abstract}

In recent years, audio-driven 3D facial animation has gained significant attention, particularly in applications such as virtual reality, gaming, and video conferencing. 
However, accurately modeling the intricate and subtle dynamics of facial expressions remains a challenge. 
Most existing studies approach the facial animation task as a single regression problem, which often fail to capture the intrinsic inter-modal relationship between speech signals and 3D facial animation and overlook their inherent consistency.
% Most existing studies approach the task of cross-modal mapping as a regression problem, which often fail to capture the intrinsic relationship between speech signals and 3D facial animation and overlook their inherent consistency.
Moreover, due to the limited availability of 3D-audio-visual datasets, approaches learning with small-size samples have poor generalizability that decreases the performance.
% . Consequently, this leads to limited generalization and difficulties in accurately representing subtle facial dynamics.
To address these issues, in this study, we propose a cross-modal dual-learning framework, termed DualTalker, aiming at improving data usage efficiency and relating cross-modal dependencies to further improve performance. The framework is trained jointly with the primary task (audio-driven facial animation) and its dual task (lip reading) and shares common audio/motion encoder components. 
% . The primary task focuses on audio-driven facial animation, while its secondary task involves lip reading. 
% These two tasks share common audio/motion encoder components, and we simultaneously train them for both primary and secondary tasks.
Our joint training framework facilitates more efficient data usage by leveraging information from both tasks and explicitly capitalizing on the complementary relationship between facial motion and audio to improve performance.
Furthermore, we introduce an auxiliary cross-modal consistency loss to mitigate the potential over-smoothing underlying the cross-modal complementary representations, enhancing the mapping of subtle facial expression dynamics.
Through extensive experiments and a perceptual user study conducted on the VOCA and BIWI datasets, we demonstrate that our approach outperforms current state-of-the-art methods both qualitatively and quantitatively. We have made our code and video demonstrations available at \url{https://github.com/Guinan-Su/DualTalker}.

\end{abstract}


\section{Introduction}
% 感觉还是有点少，但是看了其他论文都差不多这么多，后面修改时候可以再补全一些内容
% The pursuit of teaching computers to see and understand faces has provided us with insights into human behavior analysis. Audio-driven 3D facial animation has been an attractive research topic in both academia and industry and has a wide range of applications such as film production and games \cite{karras2017audio,vougioukas2020realistic,tanaka2022acceptability}. Recent investigations have made significant progress in estimating 3D face shapes, facial expressions, and facial motions from images and videos \cite{fan2022faceformer,xing2023codetalker,richard2021meshtalk}, and pushed forward the state-of-the-art significantly through deep learning. Nonetheless, these approaches potentially ignore the heterogeneity and the complex spatial-temporal discrepancy between 3D motions and speech, consequently hindering their performance. 

The quest to enable computers to perceive and comprehend faces has yielded valuable insights into human behavior analysis. One intriguing area of research that has captured the attention of both academia and industry is audio-driven 3D facial animation. This technology involves applications in various fields, including film production and gaming \cite{karras2017audio,vougioukas2020realistic,tanaka2022acceptability}. Recent advancements in this domain have led to impressive achievements in estimating 3D face shapes, facial expressions, and motions from images and videos \cite{tang2009audio,fan2022faceformer,xing2023codetalker,richard2021meshtalk}, driven by the power of deep learning. However, these approaches tend to overlook the complexities arising from the heterogeneity and intricate spatial-temporal disparities between 3D motions and speech, which can limit their overall performance.

Recent investigations can be classified into two main categories: vertex-based animation and parameter-based animation. In parameter-based animation approaches, animation curves, which are sequences of animation parameters, are generated to create smooth transitions between frames. However, these approaches tend to produce smooth animations, which can limit their performance when dealing with complex mapping rules between phoneme and viseme. On the other hand, vertex-based animation approaches directly learn mappings from audio to sequences of 3D face models by predicting mesh vertex positions. For instance, VOCA employs a time-based convolution approach to regress face movement from audio. FaceFormer uses a transformer-based autoregressive method to accommodate considerable audio information \cite{fan2022faceformer}. CodeTalker introduces discrete motion priors by training a vector quantized autoencoder (VQ-VAE) to self-reconstruct real facial movements \cite{xing2023codetalker}.

However, generating human-like motions remains a challenging task due to several factors: (1) Limited data availability: High-quality datasets with accurate 3D facial geometry and corresponding audio are scarce. When dealing with high-dimensional cross-modal features in scenarios with limited data, there's a risk of overfitting and potential performance degradation. (2) Complex facial geometry: As noted by Fan et al. \cite{fan2022joint}, the intricate geometric structure of human faces complicates the synthesis of natural facial muscle movements, which contribute to various expressions and movements. Accurately modeling these structures based on audio input is a challenging task. (3) Capturing subtle changes and nuances in audio-driven 3D face animation is nontrivial. Overlapping or anticipatory facial movements in natural speech, as well as other subtle changes, are difficult to be represented accurately. Moreover, over-smoothing in facial motion regression leads to the loss of intricate details and subtle facial dynamics when creating facial animations using regression-based methods. This technique prioritizes minimizing error, often resulting in smoothed or uniform outputs, which compromise the realism of facial animations by losing intricate details and subtle dynamics.

To address these challenges and surpass the limitations of existing research, we introduce a cross-modal dual-learning framework. This framework consists of two key components: facial animation and lip reading. For facial animation, we employ an autoregressive approach to predict a sequence of 3D face vertices, utilizing the long-term audio signal, speaker embeddings, and the historical context of facial motions. On the other hand, the lip reading component integrates speaker embeddings, face motion features, and preceding audio context to generate a sequence of audio embeddings. These two tasks are interconnected, sharing the same speaker embedding input and following an encoder-decoder pipeline with 3D face motion and audio features in reverse order. We refer to them as dual tasks due to their inherent complementary relationships. Essentially, learning to read lips can enhance the generation of 3D face animation, and vice versa, as both tasks require similar skills such as audio feature encoding, motion encoding, and the association of cross-modal information. To facilitate this dual learning, we propose a weight-sharing schema through dual cross-modal integration and sharing encoders, jointly training these two tasks.
Incorporating multiple complementary training tasks enables the model to effectively use limited data and learn more versatile and general representations. Additionally, the invertibility achieved through our parameter-sharing schemes serves as a regularization technique during the training process.

Furthermore, in the context of our dual learning framework, there still remains temporal over-smoothing in the facial animation. In this regard, we introduce an auxiliary cross-modal consistency loss that aims to capture subtle changes and nuances in audio-driven 3D facial animation. 
%Furthermore, In the context of our dual learning framework, temporal
%over-smoothing may still be present in the designed facial
%animation due to various factors, we introduce an auxiliary cross-modal consistency loss to capture subtle changes and nuances in audio-driven 3D face animation and alleviate the issue of over-smoothing in regression.
% To capture subtle changes and nuances in audio-driven 3D face animation and alleviate the issue of over-smoothing in regression, we further introduce an auxiliary cross-modal consistency loss within the dual framework. 
This approach employs a contrastive learning objective to generate semantically similar embeddings for audio and motion data. By maximizing the similarity between audio and its corresponding motion while minimizing the similarity with unrelated motion, the model effectively clusters related instances together in a high-dimensional latent space. This method captures local features and details, enhancing the model's ability to represent subtle dynamics in the data.

Two real-world datasets were included to evaluate the superiority of our method quantitatively and qualitatively. Overall, the contributions of this work can be summarized as follows:

\begin{itemize}
    \item We propose a cross-modal dual-learning framework to address limited data availability and leverage a parameter-sharing scheme for the mapping of the cross-modal complementary relationship.
    
    % . A novel parameter-sharing scheme leverages the complementary relationship between facial motion and audio, enabling versatile representation learning. 
    % \item We propose a cross-modal dual-learning framework combining facial animation and lip reading tasks to address limited data availability. A novel parameter-sharing scheme leverages the complementary relationship between facial motion and audio, enabling versatile representation learning. 
    %Feedback from the dual-task refines the primary task, improving overall performance.

    \item We employ an auxiliary cross-modal consistency loss within the dual framework to alleviate the issue of temporal over-smoothing in facial animation and to improve the model's ability to represent subtle dynamics.
    % using a contrastive learning objective to capture local features and details,
    \item Evaluated on the VOCA and BIWI datasets, our approach outperforms existing baseline methods and provides a more realistic and accurate representation of facial animations based on audio input.
\end{itemize}

The rest of our paper is structured as follows. We would like to review related methods in terms of audio-driven face animation study, dual learning, and contrastive learning in Section \ref{related works}. The details of the proposed model are introduced in Section \ref{method}. The experiment settings and provided the discussed results in Section \ref{exp}. Section \ref{con} draws the conclusions of the work.

\section{Related work}\label{related works}
\subsection{Speech-driven 3D Facial Animation}
Computer facial animation has been a long-standing endeavor, and it has garnered increasingly rapid interest in recent decades. 
Facial animation has garnered significant interest throughout the years \cite{cao2016real, fried2019text, thies2020neural, weise2011realtime, zollhofer2018state}. Although numerous 2D-based approaches have been explored \cite{lavagetto1997time, tang2009audio, wong2011audio, chen2018lip, chen2020talking, yu2020multimodal, kong2021appearance，yang2023mapping}. In our focus on this topic, we specifically emphasize 3D facial animation, distinguishing between two primary categories: vertex-based and parameter-based methods.

Regarding parameter-based 3D facial animation derived from audio, an early work by Kalberer et al. \cite{kalberer2002speech} laid the foundation. They created a viseme space using Independent Component Analysis (ICA) and employed spline fitting to interpolate sparse points mapped from phonemes to visemes within this space, resulting in smooth animations.
Taylor, et al. \cite{taylor2012dynamic} proposed an innovative approach that identified temporal units of dynamic visemes, introducing realistic motion into these units. Building on this, their subsequent work \cite{taylor2017deep} leveraged deep learning to directly learn natural coarticulation motions from data. This approach relied on an Active Appearance Model (AAM)-based parameter space.
Edwards et al. \cite{edwards2016jali} took a different route, developing a sophisticated procedural animation method based on a viseme-based rig, known as JALI. JALI enables independent control of jaw and lip actions, making it a versatile choice. Animation curves are generated based on explicit rules with empirical parameters. JALI has gained widespread adoption in the game industry due to its artist-friendly design and versatile capabilities.
A subsequent work, VisemeNet \cite{zhou2018visemenet}, introduced a deep learning approach to learn animation curves from data generated by JALI.
While parameter-based methods offer explicit control over animation, they rely on complex mapping rules between phonemes and their visual counterparts (visemes). These methods involve intricate procedures and often lack a systematic approach to animate the entire face comprehensively.

Another approach within this domain focuses on data-driven methods, with the goal of directly generating vertex-based animation from audio signals. These approaches predominantly rely on deep learning techniques.
Karras introduced a convolution-based network in their work \cite{karras2017audio}. This network incorporates a learnable emotion database and predicts vertex positions of a facial model. It operates based on 3-5 minutes of high-quality tracked mesh sequences.
Cudeiro et al. \cite{cudeiro2019capture} utilize robust audio feature extraction models to generate facial animations with various speaking styles. 
MeshTalk \cite{richard2021meshtalk} expands upon this line of research by extending the scope from lower-face to whole-face animation. It utilizes a larger dataset containing 4D scans from 250 subjects. MeshTalk successfully disentangles audio-correlated and uncorrelated facial information through a categorical latent space. However, it's important to note that the adopted latent space, while effective, may have limited expressiveness, leading to less stable animation quality when applied in data-scarce scenarios.
Additionally, FaceFormer \cite{fan2022faceformer} takes into consideration long-term audio context by incorporating transformer-based models to obtain context-related audio information. It then generates continuous facial movements in an autoregressive manner.
CodeTalker \cite{xing2023codetalker} introduces discrete motion priors by training a vector quantized autoencoder (VQ-VAE) to self-reconstruct real facial movements. This approach mitigates the problem of over-smoothing in facial animations, resulting in more realistic and nuanced facial expressions.

However, generating human-like motions remains a challenging task due to factors such as complex facial geometry and limited data availability, which make it difficult to capture the inherent consistency between speech signals and 3D facial animation. This limitation can lead to inconsistencies and a decrease in overall performance, which in turn affects the precision of lip movements and introduces ambiguity in the generated facial animations. To address these challenges, our approach employs a novel cross-modal dual-learning framework that effectively utilizes data and explicitly leverages the complementary relationship between facial motion and audio. By capitalizing on the synergy between these modalities, our innovative framework effectively addresses these issues, significantly enhancing the accuracy of lip movements.

%%%%
\subsection{Dual Learning}
Dual learning is a learning paradigm that capitalizes on the symmetrical structure present in certain learning problems, where we can interchange the input and target of one task to formulate another. This structural duality has emerged as a significant area for exploration and investigation.

In various domains, tasks have been identified with dual forms. He et al. initially introduced the concept of Dual Learning in the field of machine translation \cite{he2016dual}. Tjandra et al. extended this approach to automatic speech recognition (ASR) and text-to-speech (TTS) by treating them as dual tasks \cite{tjandra2017listening}. Tang et al. applied this idea to question answering (QA), where question generation is considered the dual task of QA. They harnessed the probabilistic correlation between QA and question generation (QG) to guide the training process \cite{tang2017question}. In computer vision, Zhu et al. employed this concept to create the image translation model CycleGAN \cite{zhu2017unpaired}.

Early work in this area focused on exploiting the duality of task pairs and proposed both supervised and unsupervised learning frameworks \cite{xia2017dual,he2016dual}. Recent studies have highlighted the significance of task duality by demonstrating that leveraging it can enhance the learning of both tasks. For instance, Su et al. used a dual supervised learning framework to simultaneously train models for natural language understanding (NLU) and natural language generation (NLG), resulting in improvements for both models \cite{su2019dual}. Shen et al. improved models for conditional text generation by drawing from computational pragmatics. They formulated language production as a game between speakers and listeners, where a speaker generates text that a listener uses to correctly identify the original input described by the text \cite{shen2019pragmatically}.

Inspired by the promising outcomes of previous research, we introduce a comprehensive learning framework that leverages the primal-dual structure of tasks for more effective data utilization. This novel learning framework capitalizes on the inherent relationships between tasks, enabling it to obtain effective feedback and regularization signals that enhance both the learning and inference processes.

\begin{figure*}
  \centering
  \includegraphics[width=1\textwidth]{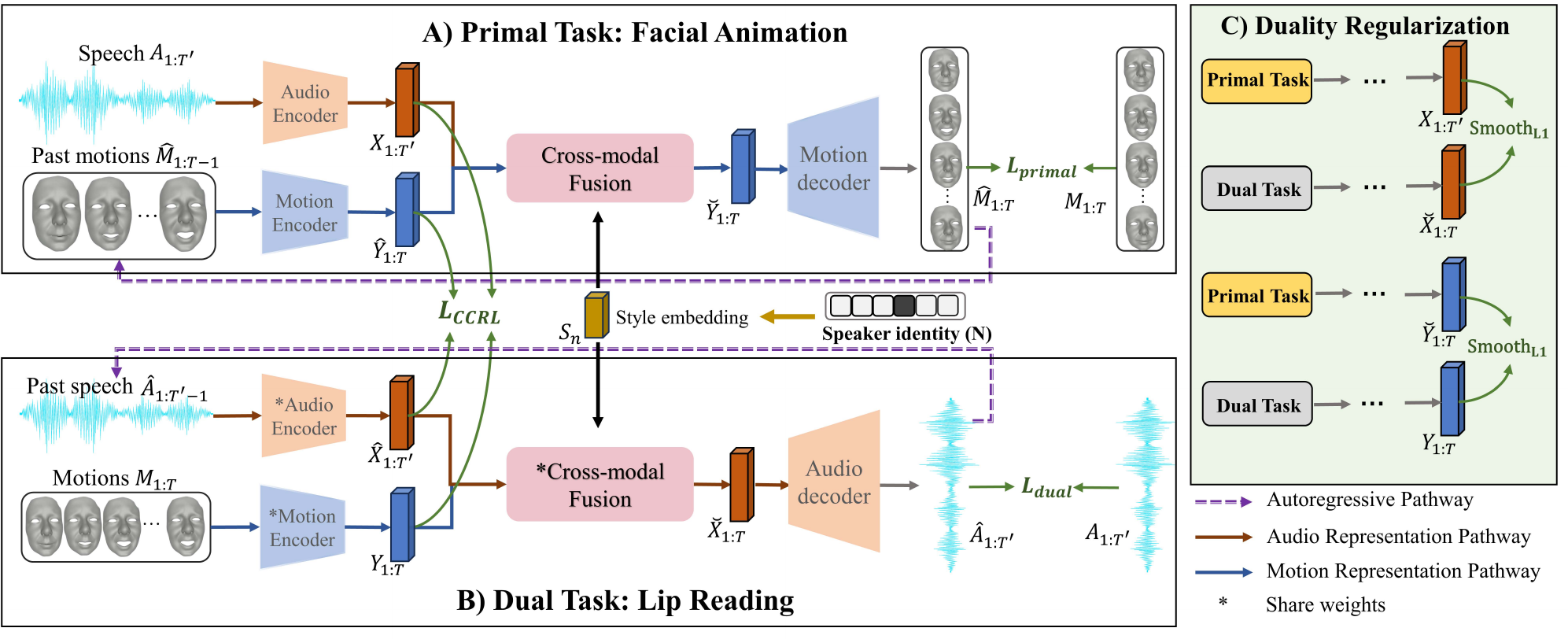}
  \caption{Overview of the proposed DualTalker in the dual learning paradigm. The network consists of two components: A) the primary task---audio-driven facial animation (top), and B) the dual task---lip reading (bottom). These two components are formulated as the inverse processes of each other. In the C) duality regularization, $\text{Smooth}_\text{{L1}}$ is leveraged to constrain the inter-modal similarity in both components. The mean squared error, ${L}_{primal}$ and ${L}_{dual}$ is applied as the loss function for these two regression problems. ${L}_{CCRL}$ denotes the auxiliary cross-modal consistency losses to mitigate over-smoothing. The two components share the audio encoder and motion encoder, where modules with * in the dual task share weights with the corresponding module in the primal task.}

  %Specifically, given an input audio signal $\mathcal{X}$, the facial animator module extracts the corresponding facial animation $\hat{m}_{T}$. The lip-reading as a dual task, on the other hand, is capable of interpreting the corresponding lip movements.}
  \label{fig1}
\end{figure*}

\subsection{Contrastive learning}

% relate the success of their
%methods to maximization of mutual information between
%latent representations 

Contrastive learning \cite{oord2018representation} \cite{hadsell2006dimensionality} is a discriminative approach that aims to bring similar samples closer together while pushing diverse samples further apart. Most self-supervised methods \cite{chen2020simple} \cite{kalantidis2020hard} \cite{kim2020adversarial} \cite{he2020momentum} \cite{xiong2020loco} \cite{bahri2021scarf} \cite{van2021revisiting} utilize data augmentation techniques, treating the corresponding augmented samples as positive pairs for the target samples and all other samples as negative pairs. In addition to single-modality representation learning, contrastive methods for multiple modalities have also been widely explored. Common methods \cite{jia2021scaling} \cite{kamath2021mdetr} \cite{zhang2022mcse} \cite{taleb2022contig} leverage cross-modal contrastive matching to align two different modalities and learn inter-modality correspondence. Beyond inter-modality contrastive approaches, Visual-Semantic Contrastive \cite{yuan2021multimodal}, XMC-GAN \cite{zhang2021cross}, CrossCLR \cite{zolfaghari2021crossclr}, GMC \cite{poklukar2022geometric}, and CrossPoint \cite{afham2022crosspoint} also introduce intra-modality contrastive learning for representation.

In comparison, supervised contrastive learning \cite{gunel2020supervised} \cite{khosla2020supervised} \cite{li2022selective} \cite{chen2022perfectly} \cite{bai2022gaussian} \cite{kang2020exploring} \cite{zeng2021modeling} considers samples from the same category as positive pairs and those from different categories as negative pairs. This approach has been shown to outperform cross-entropy loss on image classification tasks and is beneficial for learning in challenging settings, such as noisy labels, long-tailed classification, and out-of-domain detection. Supervised contrastive learning is typically used in classification problems rather than regression problems, as there is no clear definition of positive and negative pairs for continuous targets. However, a few recent papers have employed contrastive learning in the context of regression. Wang et al. \cite{wang2022contrastive} propose improving domain adaptation for gaze estimation by adding a contrastive loss term to the L1 loss. Yu et al. \cite{yu2021group} learn a model for action quality assessment from videos by contrasting two input videos and regressing their relative scores. Barbano et al. \cite{barbano2023contrastive} propose a novel contrastive learning regression loss for brain age prediction, achieving state-of-the-art performance.

Our approach aligns with these methods. We employ our auxiliary cross-modal consistency function to ensure distances in the embedding space (both inter-modal and intra-modal) are ordered according to distances in the label space. We incorporate the kernel function to define a degree of "positiveness" between samples of temporal sequences, allowing for improved alignment of motion features with audio information and alleviating the over-smoothing issue in regression.

\section{Method}\label{method}

% \begin{equation}
%     \text{Att}(\mathbf{Q^{\hat{F}},K^{\hat{F}},V^{\hat{F}}})=\text{softmax}(\frac{\mathbf{Q^{\hat{F}}(K^{\hat{F}})^T}}{\sqrt{\mathbf{d_k}}})\mathbf{V^{\hat{F}}}
% \end{equation}

% \begin{align*}
%    %\begin{equation} 
%     \text{MH}(\mathbf{Q^{\hat{F}}, K^{\hat{F}}, V^{\hat{F}} })&=\text{Concat}(\text{head}_1,...,\text{head}_{\mathbf{H}})\mathbf{W^{\hat{F}}}, \\
%     \text{where } \text{head}_h&=\text{Att}(\mathbf{Q_h^{\hat{F}},K_h^{\hat{F}},V_h^{\hat{F}}}).
% %\end{equation}
% \end{align*}

% \begin{equation}
%     \text{FDD}(M_{1:T,} \hat{M}_{1:T})=\frac{\sum_{v \in S_U}(\text{dyn}(M_{1:T}^v)-\text{dyn}(\hat{M}_{1:T}^v))}{|S_U|}
% \end{equation}

% Let $M_{1:T}=(m_1,...,m_T)$ be a sequence
% of facial motions, where each frame $m_t \in \mathbb{R}^{V \times 3}$ denotes
% the 3D movement of V vertices over a neutral-face mesh
% template $h \in \mathbb{R}^{V \times 3}$. Let further $A_{1:T}=(a_1,...,a_T)$ be
% a sequence of speech snippets, each of which $a_t \in \mathbb{R}^d$, 
% $d$ has
% $d$ samples to align with the corresponding (visual) frame
% mt. Then, our goal is to sequentially synthesize $M_{1:T}$ from
% $A_{1:T}$ so that an arbitrary neutral facial template f could be
% animated as $H_{1:T}=\{m_{1+h},...m_{T+h}\}$

\subsection{Overview}
% Cross-Modal Dual Learning
%Approach for Speech-Driven 3D Facial Animation
In this section, we introduce our proposed DualTalker, which aims at providing expressive facial animations by a cross-modal dual learning framework. The overview of DualTalker is displayed in Fig. \ref{fig1}, which consists of two components, including A) the primal task: facial animation, and B) the dual task: lip reading.

% introduce our novel cross-modal dual learning framework, termed \textbf{DualTalker}, specifically designed for speech-driven 3D facial animation.
%In this section, we present the dual learning framework of facial animation (primal task) and lip reading (dual task), \textbf{I}nvertible \textbf{A}udio-\textbf{D}riven facial animation \textbf{F}ramework (IADF).

% , A): facial animation (top) and B): lip reading (bottom).

In the facial animation component, we aim to autoregressively predict facial movements conditioned on both audio context and past facial movement sequence. Suppose that there is a sequence of 3D face movements ${M}_{1:T} = ({m}_{1}, ... , {m}_{T})$, where $T$ is the number of visual frames and each frame $m_t \in \mathbb{R}^{V \times 3}$ denotes the 3D movement of $V$ vertices over a neutral-face mesh template $h \in \mathbb{R}^{V \times 3}$. The goal here is to produce a model that can synthesize facial movements $\hat{M}_{1:T}$ that is similar to ${M}_{1:T}$ given the corresponding raw audio $\mathcal{X}$. In the encoder-decoder framework (audio encoder and motion decoder), the encoder first transforms ${A}_{1:T'} = \{{a}_{1}, {a}_{2} ..., {a}_{T^{'}}\}$ into speech representations ${X}_{1:T^{'}} = \{{x}_{1}, {x}_{2} ..., {x}_{T^{'}}\}$, where $T^{'}$ is the frame length of speech representations and ${A}_{1:T^{'}}$ is extracted speech features of $\mathcal{X}$. The style embedding layer contains a set of learnable embeddings that represents speaker identities $S = ({S}_{1}, ..., {S}_{n})$. Then, the fusion and decoder autoregressively predicts facial movements $\hat{M}_{1:T} = (\hat{m}_{1}, ..., \hat{m}_{T})$ conditioned on ${X}_{1:T^{'}}$, the style embedding ${S}_{n}$ of speaker $n$, and the past facial movements. 
Formally, we can obtain that:
\begin{equation}
   \hat{m}_{t} = \text{FacialAnimation}_{\theta}(\hat{m}_{<t}, {S}_{n}, \mathcal{X})
\end{equation}
%given a speech signal a, a Transformer-Based audio encoder is used for autoregressively obtaining the embedded audio features q ∈ Rdq , 
%a motion encoder is used for getting the historical context of motion, and MLP is used to transform the speaker v into an identity feature map.
%Then a crossmodal attention module is used to autoregressively generate a speaker-related motion feature vq ∈ Rdv from the audio feature and the identity feature. 
%Finally, a linear regression Wa is used to predict the motion for A.
where $\theta$ denotes the model parameters, $t$ is the current timestep in the sequence and $\hat{m}_{t} \in \hat{M}_{1:T}$.

The lip-reading component serves as a dual task to the facial animation component, aiming to autoregressively predict audio features based on both motion context and past audio features. We introduce a model designed to synthesize audio features  $\hat{A}_{1:T^{'}}$ that closely resemble ${A}_{1:T^{'}}$, given the motions ${M}_{1:T} = ({m}_{1}, ..., {m}_{T})$. In this dual encoder-decoder framework, which consists of a motion encoder and an audio decoder, the encoder initially transforms 3D facial movements ${M}_{1:T} = ({m}_{1}, ..., {m}_{T})$ into motion representations ${Y}_{1:T} = \{{y}_{1}, {y}_{2} ..., {y}_{T}\}$, where $T$ represents the frame length of facial motions.
%The lip reading component is a dual task of the facial animation component, which aims to autoregressively predict audio features conditioned on both motion context and past audio features. 
%A model is introduced to synthesize audio features $\hat{A}_{1:T'}$ that is similar to ${A}_{1:T'}$ given the motions ${M}_{1:T} = ({m}_{1}, ..., {m}_{T})$. In the dual encoder-decoder framework (motion encoder and audio decoder) framework, the encoder first transforms 3D face movements ${M}_{1:T} = ({m}_{1}, ..., {m}_{T})$ into motion representations ${Y}_{1:T} = \{{y}_{1}, {y}_{2} ..., {y}_{T}\}$, where $T$ is the frame length of face motions. 
%The style embedding layer contains a set of learnable embeddings that represents speaker identities $S = ({s}_{1}, ..., {s}_{n})$. 
Then, the decoder autoregressively predicts facial movements $\hat{A}_{1:T} = (\hat{a}_{1}, ..., \hat{a}_{T})$ conditioned on ${M}_{1:T}$, the style embedding ${S}_{n}$ of speaker $n$, and the past speech feature. Similarity, we can obtain:
\begin{equation}
    \hat{a}_{t} = \text{LipReading}_{\gamma}(\hat{a}_{<t}, {S}_{n}, {M}_{1:T} )
\end{equation}
where $\gamma$ denotes the model parameters, $t$ is the current timestep in the sequence and $\hat{a}_{t} \in \hat{A}_{1:T^{'}}$.

We formulate facial animation and lip reading components as inverse processes of each other by introducing a novel parameter-sharing scheme and the duality regularizer. Consequently, we jointly train a single model with two tasks. The invertibility of the model serves as a regularization term to guide the training process. %Furthermore, we employ the auxiliary cross-modal consistency loss to address the potential over-smoothing issues.
%inherent in audio-driven facial animation.

In the remainder of this section, we provide a detailed description of each component within our method's architecture, elucidating how our approach effectively addresses the challenges associated with speech-driven 3D facial animation.

\subsection{Primal Task: Audio-driven Facial Animator}

The facial animation component of our proposed DualTalker is formulated as autoregressively generating realistic and comprehensible facial expressions by given speech signals.
% Along with the speech, we further adopt a control on the talking styles as input, i.e., a style vector ${S}_{n}$, where ${n}$ is the dimension of the learned style space. 
Conditioning on the speech feature ${X}_{1:T'}$, and the style vector ${S}_{n}$, a temporal autoregressive model, composed of an audio encoder, motion encoder, and a cross-modal fusion and motion decoder, is employed to learn over the facial motion space, as depicted in Fig. \ref{fig1} A). 
%This model comprises an audio feature extractor and a multi-layer transformer encoder. 
% Audio-attentive Module, Speaker-Aware Module, and Audio-Visual Integration Module 
The audio feature extractor in our system is inspired by the state-of-the-art self-supervised pre-trained speech model, wav2vec 2.0 \cite{baevski2020wav2vec}. This component transforms raw waveform speech into feature vectors using a Temporal Convolutional Network (TCN) \cite{lea2017temporal,lea2016temporal}. Leveraging the efficient attention mechanism, the transformer-based audio feature extractor processes the audio signal to generate contextualized audio features. These features are then fed into an MLP-based (Multilayer Perceptron) audio encoder to obtain compressed audio features. Additionally, we employ an MLP-based motion encoder to derive motion features from the provided facial vertex positions.
%Following FaceFormer, our audio feature extractor adopts the architecture of the state-of-the-art self-supervised pre-trained speech model, wav2vec 2.0, which consists of an audio feature extractor and a multi-layer transformer encoder. The audio feature extractor converts the speech of raw waveform into feature vectors through a temporal convolutions network (TCN). Benefiting from the effective attention scheme, the transformer audio feature extractor converts the audio signal into the contextualized audio feature, then we fed the contextualized audio feature to an MLP-based audio encoder to get compressed audio features. We use an MLP-based motion encoder to get motion features with the given facial vertex positions.

%\includegraphics[scale=0.5]
\begin{figure}[h] %figure环境，h默认参数是可以浮动，不是固定在当前位置。如果要不浮动，你就可以使用大写float宏包的H参数，固定图片在当前位置，禁止浮动。
    \centering %使图片居中显示
    \includegraphics[width=0.5\textwidth ]{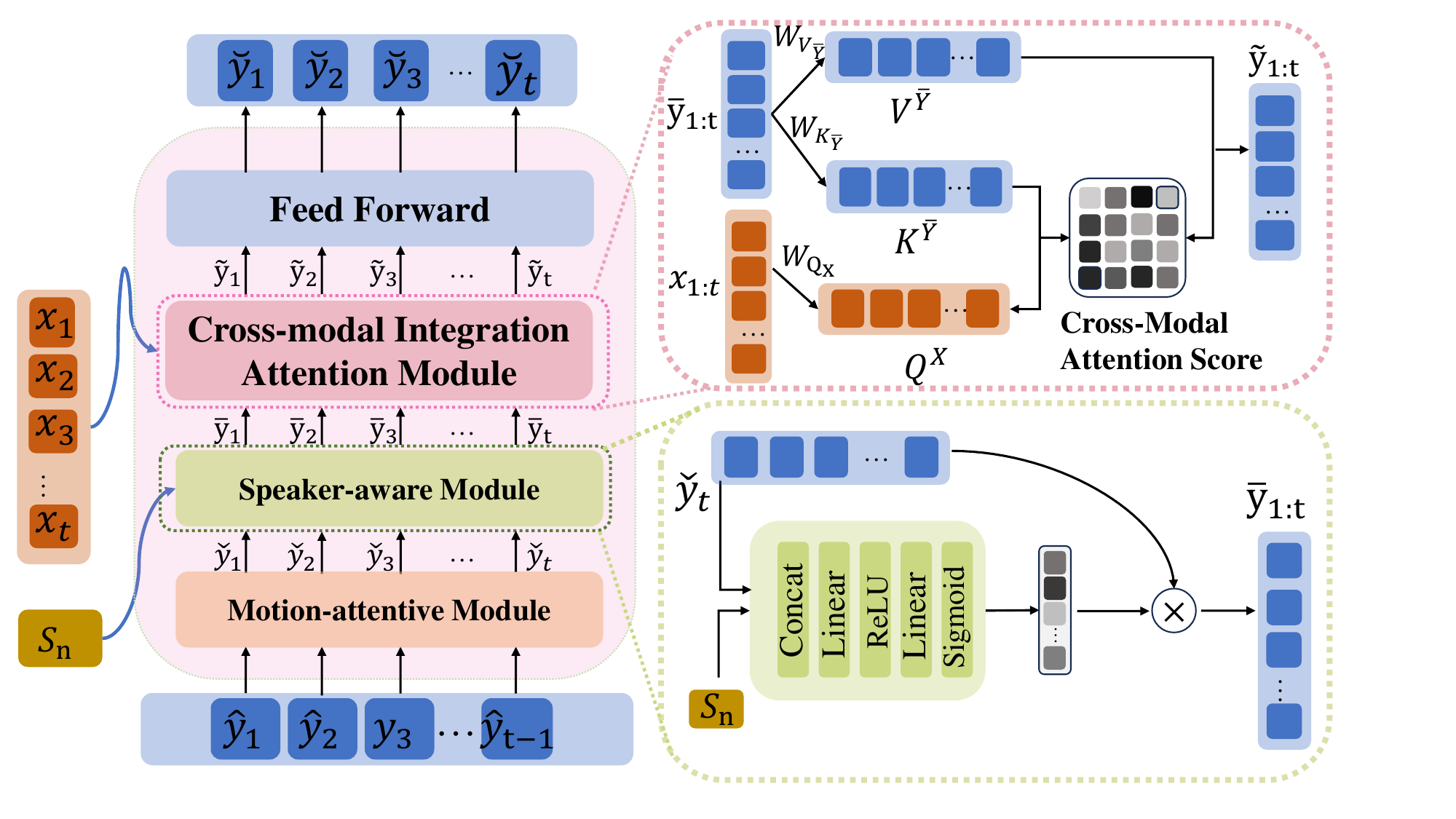} %中括号中的参数是设置图片充满文档的大小，你也可以使用小数来缩小图片的尺寸。
    \caption{The cross-modal fusion of the audio-driven facial animator} %caption是用来给图片加上图题的
    \label{fig2} %这是添加标签，方便在文章中引用图片。
\end{figure}%figure环境

%Our decoder contains multi-head self-attention, speaker-attentive module, and multi-head cross-modal attention, we equip the cross-modal decoder with multi-head self-attention to learn the dependencies between each frame in the context of the past facial motion sequence, The output intermediate features and style vector are further used to get speaker-specific features by speaker-attentive module, The output features and the past motion features and with multi-head cross-modal attention to aligning the audio and motion modalities. The output features $\hat{X}_{1:T-1}$ is further decoded by the motion decoder. The newly predicted motion $\hat{M}_{t}$ is used to update the past motions as $\hat{M}_{1:t}$, in preparation for the next prediction. 

Our cross-modal fusion consists of a motion-attentive module, speaker-aware module, and audio-visual integration module, as illustrated in Fig. \ref{fig2}. The motion-attentive module utilizes multi-head self-attention to understand the dependencies between each frame in the context of previous facial motion sequences. The resulting intermediate features and style vector are then employed by the speaker-aware module to generate speaker-specific features. These output features, combined with past motion features, are aligned using the audio-visual integration module to synchronize the audio and motion modalities. The motion decoder further processes the output features to produce the newly predicted motion. This updated motion is then used to prepare for the subsequent prediction.

%The motion decoder further processes the output features, denoted as $\breve{Y}_{1:t}$, to produce the newly predicted motion, $\hat{M}_{t}$. This updated motion, $\hat{M}_{1:t}$, is then used to prepare for the subsequent prediction.

%Formally, this recursive process can be written as

%We adapted multi-head (MH) self-attention mechanism to learn the dependencies between each frame in the context of the input sequence past facial motion sequence. 
%Given the temporally encoded facial motion representation sequence $\hat{Y}_{1:T}$,  MH self-attention first linearly projects $\hat{Y}_{1:T}$ into queries $Q^{\hat{F}}$ and keys $K^{\hat{Y}}$  of dimension $\mathbf{d_k}$, and values $V^{\hat{Y}}$ of dimension $\mathbf{d_v}$. To learn the dependencies between each frame in the context of the past facial motion sequence, a weighted contextual representation is calculated by performing the scaled dot-product attention:

Given the temporally encoded facial motion representation sequence $\hat{Y}_{1:T}$, the motion-attentive module using the multi-head (MH) self-attention first linearly projects $\hat{Y}_{1:T}$ into query vectors $Q^{\hat{Y}}$ and key vectors $K^{\hat{Y}}$ with dimensions $d_k$, as well as value vectors $V^{\hat{Y}}$ with dimensions $d_v$. To relate the dependencies between each frame within the context of previous facial motion sequences, a weighted contextual representation is computed using the scaled dot-product attention mechanism:
\begin{equation}
  \text{Att}(Q^{\hat{Y}},{K^{\hat{Y}},V^{\hat{Y}}})=\text{softmax}(\frac{Q^{\hat{Y}}({K^{\hat{Y}}})^T}{\sqrt{d_k}})V^{\hat{Y}}
\end{equation}

The MH attention mechanism, comprising parallel scaled dot-product attentions, is employed to simultaneously extract complementary information from multiple representation subspaces. The outputs from the $H$ heads are concatenated and subsequently projected forward by a parameter matrix $W^{\hat{Y}}$:

%The MH attention mechanism, which consists of $H$ parallel scaled dot-product attentions, is applied to jointly extract the complementary information from multiple representation subspaces. The outputs of $H$ heads are concatenated together and projected forward by a parameter matrix
%$W^{\hat{F}}$:

\begin{align}\label{eq4}
  %\begin{equation} 
   \text{MH}({Q^{\hat{Y}}}, {K^{\hat{Y}}}, {V^{\hat{Y}} })&=\text{Concat}(\text{head}_1,...,\text{head}_{H})W^{\hat{Y}},\\
   \text{where } \text{head}_h&=\text{Att}(Q_h^{\hat{Y}},K_h^{\hat{Y}},V_h^{\hat{Y}}). \notag
%\end{equation}
\end{align}

%Inspired by the Squeeze-and-Excitation (SE) module, we propose a speaker-attentive module to calculate the audio-speaker vector as the additional modulation parameter for audio features. Fig. \ref{fig2} illustrates our attention mechanism. It enables the network to adjust the degrees of contributions from audio feature input, depending on the content of the style embeddings. Initially, the module concatenates the two feature maps style embeddings and audio features along the channel dimension and then fed to a multi-layer perceptron (MLP) \text{H} to obtain Sigmoid-normalized fusion weights for the weighted audio feature and the MLP comprises a FC layer with $\frac{2\times{\mathbf{d_k}}}{r}$ output channels (r is the channel compression ratio), ReLU function, a FC layer, and Sigmoid function. 
%\breve

Inspired by the Squeeze-and-Excitation (SE) module\cite{hu2018squeeze}, we propose a speaker-aware module to compute the audio-speaker vector, which serves as an additional modulation parameter for audio features. The attention is depicted in the lower right corner of Fig. \ref{fig2}, which allows the network to adjust the contributions from audio feature input based on the content of style embeddings. Initially, the module concatenates the style embeddings and audio features along the channel dimension. This combined data is then fed into an MLP to obtain sigmoid-normalized fusion weights for the weighted audio feature.

The MLP consists of a fully connected (FC) layer with $\frac{2\times{d_k}}{r}$ (${r}$ is the channel compression ratio) output channels a ReLU function, another FC layer, and a Sigmoid function.
Formally, we can obtain that:
\begin{equation}
\bar{\text{Y}}_{1:T}= \text{MLP}(\text{Concat}(\text{S}_{n},\breve{\text{Y}}_{1:T})) \cdot \breve{\text{Y}}_{1:T}
%(\frac{\mathbf{Q^{\hat{F}}(K^{\hat{F}})^T}}{\sqrt{\mathbf{d_k}}})\mathbf{V^{\hat{F}}}
\end{equation}

%The cross-modal multi-head attention aims to combine the outputs of the audio encoder (speech features) and speaker-attentive module (speaker-specific motion features) to align the audio and motion modalities (see Fig. \ref{fig2}). Both $\bar{Y}_{1:T}$ and ${X}_{1:T}$ are fed into  cross-modal MH attention. Likewise, $\bar{Y}_{1:T}$ is transformed into two separate matrices: keys $K^{\bar{Y}}$ and values $\mathbf{V^{\bar{Y}}}$, whereas ${X}_{1:T}$ is transformed into queries $\mathbf{Q^{{X}}}$. The output is calculated as a weighted sum of $\mathbf{V^{\bar{Y}}}$,

The audio-visual integration module aims to integrate the outputs from the audio encoder (speech features) and the speaker-aware module (speaker-specific motion features) to align the audio and motion modalities. Both $\bar{Y}_{1:T}$ and ${X}_{1:T}$ are inputted into the cross-modal MH attention. Subsequently, $\bar{Y}_{1:T}$ is transformed into two distinct matrices: keys $K^{\bar{Y}}$ and values $V^{\bar{Y}}$, while ${X}_{1:T}$  is converted into query vectors $Q^{{X}}$ .The output is calculated as a weighted sum of $V^{\bar{Y}}$, using the following equation:

\begin{equation}
  \text{Att}(Q^{{X}},K^{\bar{Y}},V^{\bar{Y}})=\text{softmax}(\frac{Q^{{X}}(K^{\bar{Y}})^T}{\sqrt{d_k}})V^{\bar{Y}}
\end{equation}

%To explore different subspaces, we also extend to H heads as in Eq. \ref{eq4}. Finally, the predicted face motion $\hat{M}_{1:T}$ is obtained by projecting the d-dimensional hidden state back to the V-dimensional 3D vertex space via a motion decoder.
To investigate various subspaces, we also extend the approach to $H$ heads, as shown in Eq. \ref{eq4}. Finally, the predicted facial motion $\hat{M}_{1:T}$ is obtained by projecting the hidden state back into the $V$-dimensional 3D vertex space using a motion decoder.

\subsection{Dual Task: Lip reading}
%Lip reading component of our proposed IADF is formulated as a task generating an audio feature sequence given a facial motion and a specific speaker. It's a dual task of audio-driven Facial Animation. 
%During training, our target is to learn a model such that the generated audio feature embedding is similar to the referenced one. 
%The generation of each frame of the audio feature can be written as:
%Lip reading shares the audio encoder, motion encoder, and style encoder with its primal task: facial animation. The cross-modal attention module is also utilized for the motion self-attention module, speaker-attentive module, and audio $\and$ motion representations fusion at B. 

% Motion-Focused Module
%Speaker-Aware Module
%Audio-Visual Integration Module
The lip-reading component of our proposed DualTalker is designed as a task that generates an audio feature sequence given facial motion and a specific speaker, serving as a dual task to audio-driven facial animation. It shares the audio encoder, motion encoder, and style encoder with its primal task. The cross-modal fusion module is similarly utilized for the audio-attentive module, speaker-aware module, and audio-visual integration module in the lip-reading component.

\begin{figure}[h] %figure环境，h默认参数是可以浮动，不是固定在当前位置。如果要不浮动，你就可以使用大写float宏包的H参数，固定图片在当前位置，禁止浮动。
    \centering %使图片居中显示
    \includegraphics[width=0.4\textwidth]{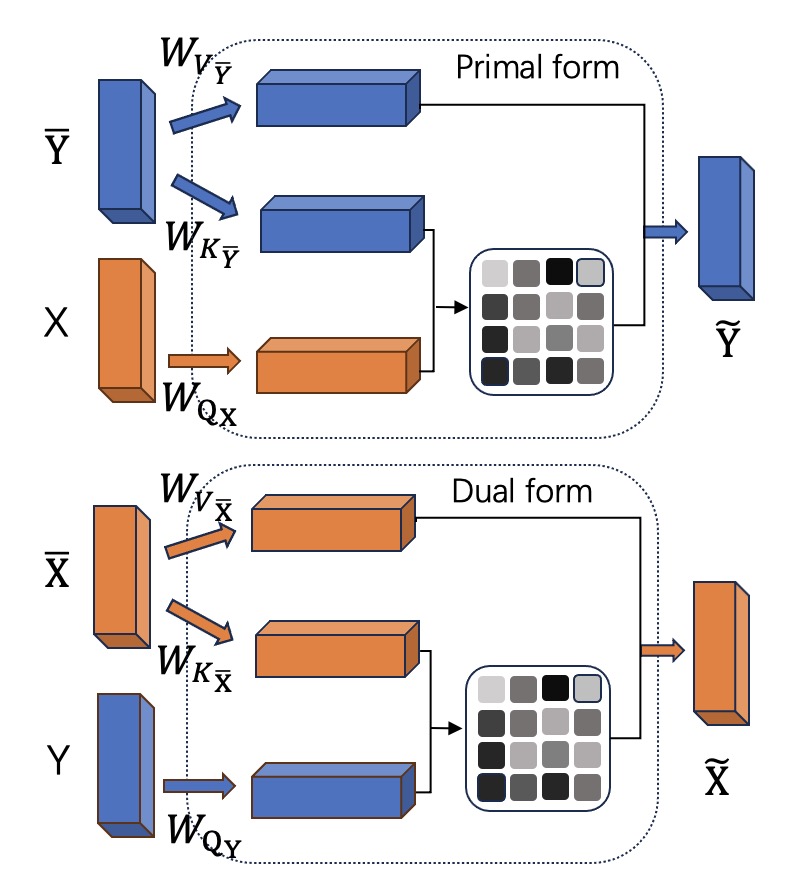} %中括号中的参数是设置图片充满文档的大小，你也可以使用小数来缩小图片的尺寸。
    \caption{The dual form of cross-modal integration module} %caption是用来给图片加上图题的
    \label{fig3} %这是添加标签，方便在文章中引用图片。
\end{figure}%figure环境

%Similar to Eq., the inference of audio features can be formulated as:
%The B cross-modal attention module could be viewed as the dual form of the A (as in Fig. \ref{fig3}).  It also aims to align the audio and motion modalities. Likewise, The input $\bar{X}_{1:T}$ is transformed into two separate matrices: keys $K^{\bar{X}}$ and values $\mathbf{V^{\bar{X}}}$, whereas ${Y}_{1:T}$ is transformed into queries $\mathbf{Q^{{Y}}}$. The output is calculated as a weighted sum of $\mathbf{V^{\bar{X}}}$. As shown in Fig. \ref{fig3}, we unify the two dual modules by sharing parameters  $W_{Q_{X}} = W_{K_{\bar{X}}}$ and ${W}_{{Q}_{Y}} = {W}_{{K}_{\bar{Y}}}$.
%${W}_{{Q}_{X}} = {W}_{{K}_\bar{X}}$
%multi-head self-attention, a speaker-attentive module, and multi-head cross-modal attention

Specifically, the audio-visual integration module in lip reading can be considered as the dual form of visual-audio integration module in facial animation (as shown in Fig. \ref{fig3}) and also aims to align the audio and motion modalities. Similarly, the input $\bar{X}_{1:T^{'}}$ is transformed into two separate matrices: keys $K^{\bar{X}}$ and values $V^{\bar{X}}$, while ${Y}_{1:T}$ is converted into query vectors $Q^{{Y}}$. The output is calculated as a weighted sum of $V^{\bar{X}}$. As depicted in Fig. \ref{fig3}, we unify the two dual modules by sharing parameters $W_{Q_{X}} = W_{K_{\bar{X}}}$ and ${W}_{{Q}_{Y}} = {W}_{{K}_{\bar{Y}}}$.

\begin{equation}
  \text{Att}(Q^{{Y}},K^{\bar{X}},V^{\bar{X}})=\text{softmax}(\frac{Q^{\bar{Y}}(K^{\bar{X}})^T}{\sqrt{d_k}})V^{\bar{X}}
\end{equation}

Finally, the predicted audio features $\Breve{X}_{1:T'}$ are fed into an MLP-based audio decoder to generate the predicted audio embeddings.
%the B crossmodal attention module could be viewed as the dual form of the A. 
%In the next section, we will introduce our attempt to investigate the connection between the two crossmodal attention modules.
\subsection{Duality Regularization Loss}
%With the Dual learning framework, we have formulated facial animation and lip reading as the inverse process to each other. Consequently, given an audio/motion pair (${X}_{1:T}$ , ${Y}_{1:T}$ ), the predicted fusion feature of MH cross-modal attention: motion/audio representations ($\Breve{X}_{1:T}$ , $\Breve{Y}_{1:T}$ ) are expected to unify.
In the context of our dual learning framework, we have designed facial animation and lip reading as inverse processes of each other. Consequently, when given an audio/motion pair (${X}_{1:T^{'}}$, ${Y}_{1:T}$), the predicted fusion feature of cross-modal fusion, which includes motion/audio representations ($\Breve{X}_{1:T^{'}}$, $\Breve{Y}_{1:T}$), is expected to converge:

\begin{equation}
%\begin{align}
{L}_\text{DR} = \text{Smooth}_{\text{L1}}({X}_{1:T^{'}}, \Breve{X}_{1:T^{'}}  )  + \text{Smooth}_{\text{L1}}({Y}_{1:T}, \Breve{Y}_{1:T}  )
%\end{align}
\end{equation}

By minimizing the smooth L1 loss between the motion/audio representations, facial animation and lip reading are interconnected. Furthermore, the duality regularizer can be considered as an approach to provide soft targets for motion/audio feature learning, ensuring coherent semantic expression and accurate content delivery.
%by minimizing the smooth L1 loss between the motion/audio representations,  facial animation and lip reading are linked with each other. Moreover, the Duality Regularizer can be viewed as a way to provide soft targets for the motion/audio feature learning.

\subsection{Cross-modality Consistency Regularization Loss}

In the context of our dual learning framework, temporal over-smoothing may still exist in the designed facial animation. The regression-to-mean problem, which occurs when treating audio-driven facial animation as a regression task, results in over-smoothed facial motions. Furthermore, the data's inherent ambiguity arises from aspects such as co-articulation, where adjacent phonemes influence each other, leading to different articulatory movements and multiple potential facial animations. Non-verbal cues, including facial expressions like raised eyebrows, smiles, or frowns, also contribute to this ambiguity by introducing variations in facial animations.

To address these challenges, we propose an auxiliary cross-modal consistency loss to improve subtle facial expression dynamics. We take into account similarities and differences among intra-/inter-modal instances by placing similar instances close together while distancing dissimilar instances The main advantage of our method is its ability to capture local features and details in the data, thereby enhancing model performance in overcoming temporal over-smoothing problems.

In our method, the input modalities (${X}_{1:T'}$, ${Y}_{1:T}$) and ($\hat{X}_{1:T'}$, $\hat{Y}_{1:T}$) have already been processed by encoder networks. Our objective is to keep the semantically related inputs close, both in the original feature embedding (intra-modality) and in the joint embedding space (inter-modality).

To elaborate, let's take (${X}_{1:T'}$, ${Y}_{1:T}$) as an example. We employ a similarity function, Sim (e.g., cosine similarity), to measure inter-modal  $s_t^\text{inter} = \text{Sim}(X_k, Y_t)$  and intra-modal similarities $s_t^\text{intra} = \text{Sim}(X_k, X_t)$  for a given $k$-th sample $X_k$ in the sequence. Here, $X_t$ and $Y_t$ represent the $t$-th sample of ${X}_{1:T}$/${Y}_{1:T^{'}}$. In this study, we implement contrastive learning to map inter/intra-modal similarities by treating $(X_k, Y_k)$ and $(X_k, X_k)$ as positive samples and others ($(X_k, X_t/Y_t), t\neq k$) as negative samples. Furthermore, we suggest using a kernel function ${w}_{t}= \text{K}({M}_{k} - {M}_{t})$ to measure the distance between positive and negative samples in a weighted manner. Here, $\text{K}(\cdot)$ denotes a Gaussian kernel function (RBF) that maps the weights into the range $(0,1)$. Our goal is to learn maps samples with a high degree of positiveness  ($w_{t} \sim 1$) close in the latent space and samples with a low degree ($w_{t} \sim 0$) far apart. Finally, we propose a formulation $L_\text{CCRL}$ to enhance the repulsion strength for samples in proportion to their distance from the anchor in the kernel space.

\begin{equation}
\begin{gathered}
   w_k\left[s_t^\text{intra}\left(1-w_t\right)-s_k^\text{intra}\right] \leq 0, \quad \forall k, t \neq k \in T(k) \\
   w_k\left[s_t^\text{inter}\left(1-w_t\right)-s_k^\text{inter}\right] \leq 0, \quad \forall k, t \neq k \in T(k) \notag
\end{gathered}
\end{equation}

\begin{gather}
       \mathcal{L}_\text{CCRL}({X}_{1:T'}, {Y}_{1:T})=-\frac{1}{\sum_t w_t} \sum_k w_k \log \frac{\exp \left(s_k^\text{inter}\right)}{s^{'}} \\
s^{'} = \sum_{t \neq k} \exp \left(s_t^\text{inter}\left(1-w_t\right)\right) + \exp \left(s_t^\text{intra}\left(1-w_t\right)\right)
\end{gather}

We use $T(k)$ to represent the indices of samples in the current timestamp $T$. The weighting factor $1-{w}_{t}$ acts as a temperature value, assigning more weight to samples further away from the anchor in the kernel space. With an appropriate kernel choice, samples closer to the anchor will experience very low repulsion strength $(\sim 0)$. We believe that this approach is better suited for continuous attributes (i.e., regression tasks) as it ensures that samples close in the kernel space will also be close in the representation space.

\begin{gather}
       \mathcal{L}_\text{CCRL}= {L}_\text{CCRL}({X}_{1:T'}, {Y}_{1:T}) + {L}_\text{CCRL}(\hat{X}_{1:T'}, \hat{Y}_{1:T})
\end{gather}

We employ this auxiliary cross-modal consistency loss to bring the features (${X}_{1:T'}$, ${Y}_{1:T}$) and ($\hat{X}_{1:T'}$, $\hat{Y}_{1:T}$) of semantically similar input examples closer together. This process adjusts motion features according to audio features through back-propagation and vice versa. 
% Specifically, our method aims to enhance the overall performance of the model by addressing over-smoothing issues and preserving the proximity of semantically related inputs.

\subsection{Dual training}
With our proposed weight-sharing schema (dual cross-modal integration and sharing encoders), our two components can be reconstructed as the inverse processes of each other. Hence, joint training on these two tasks introduces the invertibility of the model as an additional regularization term to guide the training process. The overall training loss is as follows: 
\begin{align*}
    Loss = &\lambda_1 {L}_\text{Primal}({M}_{1:T}, \Breve{M}_{1:T}) + \lambda_2 {L}_\text{Dual}({A}_{1:T^{'}}, \Breve{A}_{1:T^{'}}) + \\ &\lambda_3 
    {L}_\text{DR} + \lambda_4 
 {L}_\text{CCRL}
\end{align*}
where ${L}_\text{Primal}(M, \Breve{M})$ and ${L}_\text{Dual}(A, \Breve{A})$ adopt the regression loss (We use the Mean Squared Error, MSE, as the loss function here) as the training loss for facial animation and lip reading, respectively. Finally, since every operation is differentiable, the entire model can be trained in an end-to-end manner. We formulate the loss function with the weights $\lambda_1$, $\lambda_2$, $\lambda_3$, and $\lambda_4$ for balancing the training loss.

% In the next section, we will demonstrate that our dual training strategy can significantly improve the performance of the audio-driven facial animator, ensuring coherent semantic expression and accurate content representation.

\section{experience and results}\label{exp}

\begin{figure*}
  \centering
  \includegraphics[width=1\textwidth]{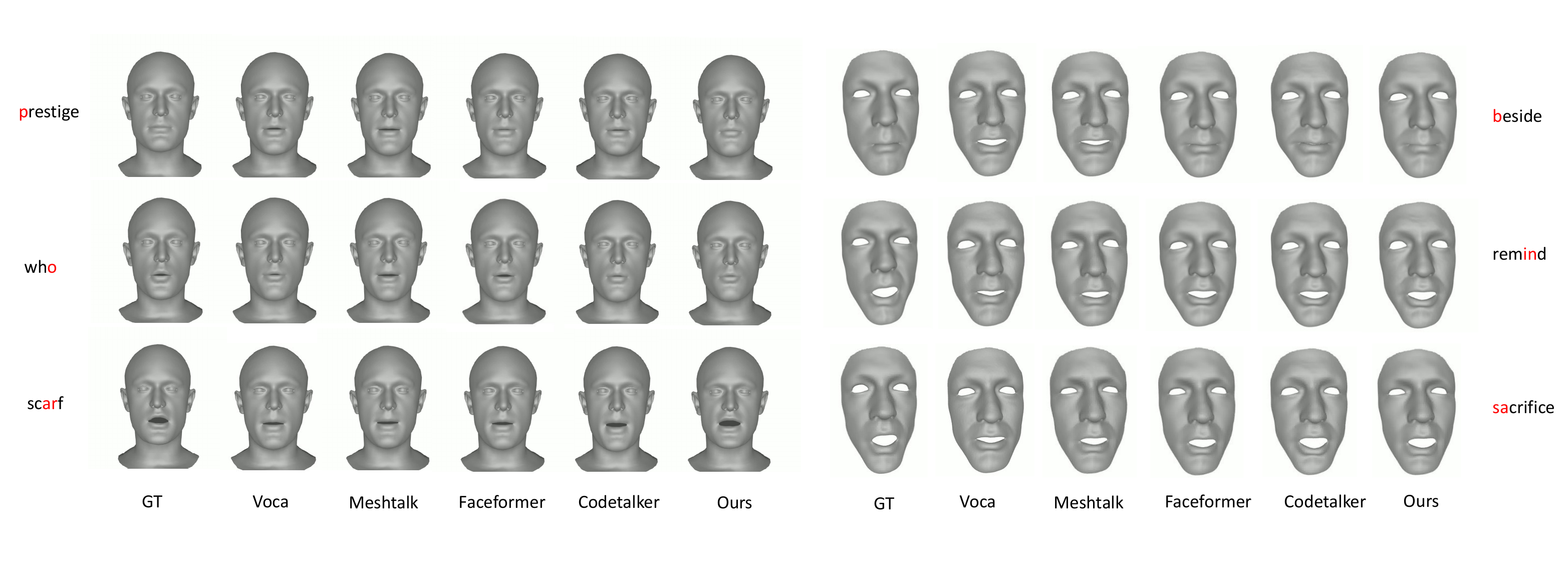}
  \caption{Visual comparisons of sampled facial motions animated by different methods on VOCA-Test (left) and BIWI-Test-B (right).}
  \label{fig4}
\end{figure*}
\subsection{Datasets} We carried out training and testing using two publicly available 3D datasets: BIWI \cite{fanelli20103} and VOCASET \cite{cudeiro2019capture}. Both datasets include audio-3D scan pairs that exhibit English speech pronunciation. VOCASET features 255 unique sentences, some of which are shared across speakers. In contrast, BIWI comprises 40 unique sentences shared among all speakers, making the task more challenging due to the limited phonemic information provided.

%\textbf{VOCASET Dataset}. VOCASET contains 480 facial motion sequences from 12 subjects, captured at 60fps for approximately 4 seconds per sequence. Each 3D face mesh is registered to the FLAME \cite{li2017learning} topology with 5023 vertices. We adopted the same training (VOCA-Train), validation (VOCA-Val), and testing (VOCA-Test) splits as FaceFormer \cite{fan2022faceformer} and CodeTalker\cite{xing2023codetalker} for fair comparisons.

\textbf{VOCASET Dataset}. VOCASET contains 480 facial motion sequences from 12 subjects, captured at 60fps for approximately 4 seconds per sequence. Each 3D face mesh is registered to the FLAME \cite{li2017learning} topology with 5023 vertices. We adopted the same training (VOCA-Train), validation (VOCA-Val), and testing (VOCA-Test) splits as FaceFormer \cite{fan2022faceformer} and CodeTalker\cite{xing2023codetalker} to ensure fair comparisons.

\textbf{BIWI Dataset}. BIWI is a 3D audio-visual corpus that showcases emotional speech and facial expressions in dynamic 3D face geometry. It consists of two parts: one with emotions and the other without, featuring 40 sentences spoken by 14 subjects – six males and eight females. Each recording is repeated twice in neutral or emotional situations, capturing dynamic 3D faces at a 25fps rate. The registered topology exhibits 23370 vertices, with an average sequence length of around 4.67 seconds. We followed the data split from FaceFormer \cite{fan2022faceformer} and CodeTalker\cite{xing2023codetalker}, using only the emotional subset. Specifically, the training set (BIWI-Train) contains 190 sentences (two facial data were officially lost), while the validation set (BIWI-Val) includes 24 sentences. There are two testing sets: BIWI-Test-A comprises 24 sentences spoken by 6 participants seen during the training process, suitable for both quantitative and qualitative evaluations. In contrast, BIWI-Test-B contains 32 sentences spoken by eight unseen participants, making it more appropriate for qualitative evaluation.

\subsection{Baseline Implementations}
We conducted experiments on the BIWI and VOCASET datasets to compare the performance of our proposed model with state-of-the-art methods, including VOCA \cite{cudeiro2019capture}, MeshTalk \cite{richard2021meshtalk}, FaceFormer \cite{fan2022faceformer}, and CodeTalker \cite{xing2023codetalker}. We trained and tested VOCA on the BIWI dataset using its official codebase and directly tested the released model trained on VOCASET. Additionally, we implemented MeshTalk on both BIWI and VOCASET. For FaceFormer and CodeTalker, we utilized the provided pre-trained weights for testing.
\subsection{Quantitative Evaluation} 
\begin{table}[h!]
\caption{\label{demo-table}Quantitative evaluation results on BIWI-Test-A.}
\centering
  \renewcommand\arraystretch{1.2}
 \begin{tabular}{c c c } 
 \hline
 \hline
 Methods & \makecell[c]{Lip Vertex Error $\downarrow$ \\ $(×{10}^{-4} mm)$} & \makecell[c]{FDD $\downarrow$  \\ $(×{10}^{-5}mm)$ }\\ [0.5ex] 
 \hline
VOCA \cite{cudeiro2019capture} (2019) &  6.5563  & 8.1816  \\ 
MeshTalk \cite{richard2021meshtalk} (2021) & 5.9181 & 5.1025  \\
FaceFormer \cite{fan2022faceformer} (2022) & 5.3077 & 4.6408 \\
CodeTalker \cite{xing2023codetalker} (2023) & 4.7914 & 4.1170  \\
Ours & 4.6210 & 3.7424 \\ [1ex] 
 \hline
 \hline
  \label{table1}
 \end{tabular}
\end{table}

%。Ours & 4.6354 & 3.9448
To evaluate lip synchronization, we adopt the lip vertex error metric used in MeshTalk \cite{richard2021meshtalk} and FaceFormer \cite{fan2022faceformer}. This metric measures the lip deviation in a sequence concerning the ground truth, i.e., calculating the maximal L2 error of all lip vertices for each frame and taking the average over all frames. However, the lip vertex error alone may not fully reflect the lip readability of the generated facial animation.

Following the principles of CodeTalker \cite{xing2023codetalker}, we utilize the upper-face dynamics deviation metric. Facial expressions during speech are loosely correlated with the content and style of verbal communication, and as such, they vary significantly among individuals. The facial dynamics deviation (FDD) technique measures the dynamic changes in the face during a sequence of motions and compares them to the ground truth.
Specifically, the upper-face dynamics deviation (FDD) is calculated by: 
\begin{equation}
    \text{FDD}(M_{1:T,} \hat{M}_{1:T})=\frac{\sum_{v \in S_U}(\text{dyn}(M_{1:T}^v)-\text{dyn}(\hat{M}_{1:T}^v))}{|S_U|}
\end{equation}

$M^v \in \mathbb{R}^{3 \times T}$ denotes the motions of the $v$-th vertex, and $S_U$ is the index set of upper-face vertices. $\text{dyn}(\cdot)$ denotes the standard deviation of the element-wise L2 norm along the temporal axis.

We calculate the lip vertex error and upper-face dynamics deviation (FDD) over all sequences in BIWI-Test-A and take the average for comparison. According to Table \ref{table1}, the proposed DualTalker achieves lower error than the existing state-of-the-art, suggesting that it produces more accurate lip-synchronized movements. Besides, Table \ref{table1} shows that our method achieves the best performance in terms of FDD. It indicates the high consistency between the predicted upper-face expressions together with the trend of facial dynamics (conditioned on the speech and talking styles).

\subsection{Qualitative Evaluation}
%While metrics are essential for evaluating 3D talking faces, visualizing the prediction results is necessary for a more comprehensive understanding of the model’s performance. We visually compare our method with other competitors in Figure 4. For a fair comparison, we assign the same talking style to VOCA, FaceFormer, MeshTalker, CodeTalker, and our method as conditional input, which is sampled at random. To check the lip synchronization performance, we illustrate three typical frames of synthesized facial animations that speak at specific syllables, as compared to the upper partition in Figure 4. We can observe that compared with the competitors, the lip movements produced by our method are more accurately articulated with the speech signals and also more consistent with those of the Ground Truth. For example, our method produces better lip sync with proper mouth closures when pronouncing bilabial consonant /b/ (i.e., “bedside” in the upper-right case of Figure 4), compared to the other methods; for the even challenging speech parts “remind” and “sacrifice” that need to pout, Our method can produce accurate lip shapes while other methods suffer from the over-smoothing problem and failure to lip-sync correctly (Zoom in for better inspection).
While metrics play a vital role in evaluating 3D talking faces, visualizing prediction results is necessary for a more comprehensive understanding of the model's performance. We visually compare our method with other competitors in Fig. \ref{fig4}. To ensure a fair comparison, we assign the same talking style to VOCA, FaceFormer, MeshTalker, CodeTalker, and our method as conditional input, which is randomly sampled. To examine lip synchronization performance, we illustrate three typical frames of synthesized facial animations that speak at specific syllables, as shown in the upper partition of Fig. \ref{fig4}. We can observe that, compared to the competitors, the lip movements produced by our method are more accurately articulated with the speech signals and more consistent with the ground truth.

For instance, our method achieves better lip sync with proper mouth closures when pronouncing the bilabial consonant /b/ (e.g., "bedside" in the upper-right case of Fig. \ref{fig4}) compared to other methods. For even more challenging speech parts like "remind" and "sacrifice" that require pouting, our method can produce accurate lip shapes, while other methods suffer from over-smoothing problems and fail to lip-sync correctly. For a more detailed inspection, please zoom in on the visualizations. 
\subsection{User Study}

\begin{table}[h!]
\centering
\setlength\tabcolsep{12pt}
  \renewcommand\arraystretch{1.2}
\caption{User study results on BIWI-Test-B}

 \begin{tabular}{c c c } 
 \hline
 \hline
 Ours vs. Competitor & Realism & Lip Sync  \\ [0.5ex] 
 \hline
Ours vs. VOCA \cite{cudeiro2019capture} (2019) & 86.7 & 78.3  \\ 
Ours vs. MeshTalk \cite{richard2021meshtalk} (2021) & 72.1 & 71.5  \\
Ours vs. FaceFormer \cite{fan2022faceformer} (2022) & 62.5 & 61.5  \\
Ours vs. CodeTalker \cite{xing2023codetalker} (2023) & 57.3 & 52.1  \\
Ours vs. GT & 46.3 & 44.8 \\ [1ex]
 \hline
 \hline
 \end{tabular}\label{table2}
\end{table}
%  66.7 & 58.3 
% 62.1 & 61.5
\begin{table}[h!]
\caption{User study results on VOCA-Test.}
\setlength\tabcolsep{12pt}
\centering
  \renewcommand\arraystretch{1.2}
 \begin{tabular}{c c c } 
 \hline
 \hline
 Ours vs. Competitor & Realism & Lip Sync  \\ [0.5ex] 
 \hline
Ours vs. VOCA \cite{cudeiro2019capture} (2019) & 75.4 & 77.1  \\ 
Ours vs. MeshTalk \cite{richard2021meshtalk} (2021) & 76.4 & 73.4  \\
Ours vs. FaceFormer \cite{fan2022faceformer} (2022) & 61.8 & 66.7  \\
Ours vs. CodeTalker \cite{xing2023codetalker} (2023) & 63.3 & 61.5  \\
Ours vs. GT & 47.8 & 41.1 \\ [1ex]
 \hline
 \hline
 \end{tabular}\label{table3}
\end{table}
%VOCA & 65.4 & 67.1  \\ 
%Ours vs. MeshTalk & 66.4 & 63.4  \\
The human perception system has been fine-tuned over time to understand subtle facial motions and capture lip synchronization, making it the most dependable measure for speech-driven facial animation tasks. To evaluate two metrics, perceptual lip synchronization, and facial realism, we conducted a user study comparing our DualTalker method with VOCA, MeshTalk, FaceFormer, Codetalker, and the ground truth. We employed A/B tests for each comparison, specifically ours vs. our competitor's, in terms of realistic facial animation and lip sync. For BIWI, we randomly selected 30 samples from BIWI-Test-B for each of the five comparison types. To achieve the greatest variation in speaking styles, we ensured the sampling results fairly encompassed all conditioning styles. As a result, 180 A vs. B pairs (30 samples and 6 comparisons) were created for BIWI-Test-B. Each pair was evaluated by at least three different participants separately. For the user study on VOCASET, we applied the same settings as those on the BIWI dataset, i.e., another 180 A vs. B pairs from the VOCA-Test set. A total of 30 participants with excellent vision and hearing abilities successfully completed the evaluation. Moreover, each participant engaged in all eight types of comparisons to ensure a wider exposure and to accommodate the diverse range of preferences.

The A/B testing results in terms of lip sync and realism on BIWI-Test-B are shown in Table \ref{table2}, indicating that participants favored our method over the competitors. This preference can be ascribed to the facial animations produced by our method, which exhibit more expressive facial motions, accurate lip shapes, and well-synchronized mouth movements. Even for the VOCA-Test, which inherently possesses fewer upper-face motions, a similar preference for our method can be observed in Table \ref{table3}. The dual-learning framework employed in our method allows for enhanced semantic understanding and greater natural realism. In conclusion, the user study confirms that the facial animations generated by our method display exceptional perceptual quality.

\subsection{Ablation Studies}

% Distance between lower and upper lip within a sampled
%sequence from VOCA-Test of (a) reconstruction

\begin{figure}[h] %figure环境，h默认参数是可以浮动，不是固定在当前位置。如果要不浮动，你就可以使用大写float宏包的H参数，固定图片在当前位置，禁止浮动。
    \centering %使图片居中显示
    \includegraphics[width=0.5\textwidth]{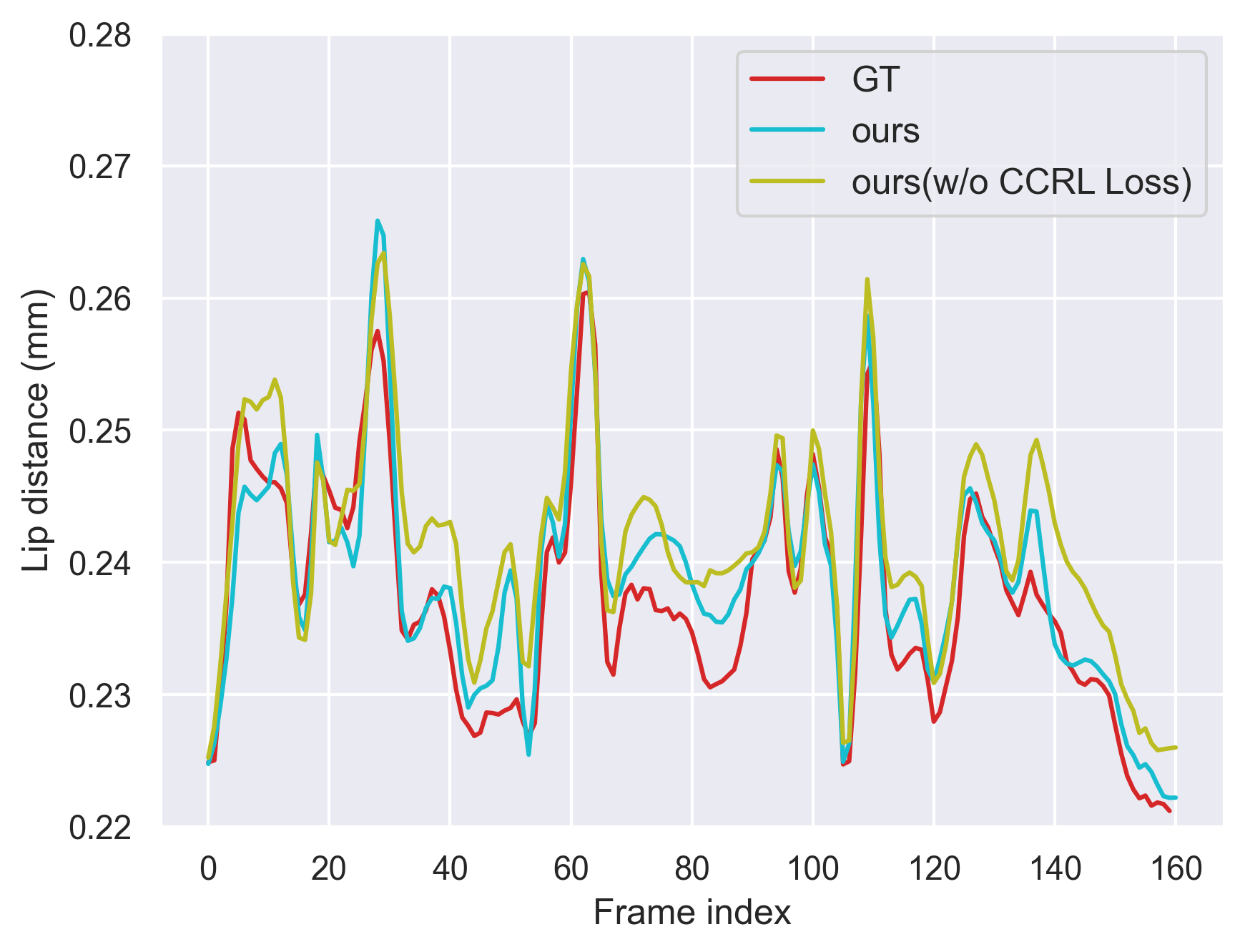} %中括号中的参数是设置图片充满文档的大小，你也可以使用小数来缩小图片的尺寸。
    \caption{ Distance between lower and upper lip within a sampled sequence from VOCA-Test-A} %caption是用来给图片加上图题的
    \label{fig5} %这是添加标签，方便在文章中引用图片。
\end{figure}%figure环境

\begin{table}[h!]
\caption{\label{demo-table}Ablation study for our components on BIWI-Test-A.}
\centering
   \setlength\tabcolsep{6pt}
  \renewcommand\arraystretch{1.4}
 \begin{tabular}{c c c } 
 \hline
 \hline
 Methods & \makecell{Lip Vertex Error $\downarrow$ \\ $(×{10}^{-4} mm)$} & \makecell{FDD $\downarrow$ \\ $(×{10}^{-5}mm)$}\\ [0.5ex] 
 \hline
\makecell{w/o Cross-modality \\Consistency Regularization Loss } & 4.6354 & 3.9448 \\
%\makecell{w/o Dual Training \\Diagram} & 5.3077 & 4.6408 \\
\makecell{w/o Dual Training \\Diagram} & 5.2841 & 4.4112 \\
\makecell{DeepSpeech encoding} & 6.0915 & 5.8401 \\
%our & 4.79 & 3.8 \\ [1ex]   Ours &。4.6354 & 3.9448
 \hline
 \hline
 \end{tabular}\label{table4}

\end{table}

\textbf{Impact of the audio feature extraction:} We discovered that Wav2vec 2.0 excels in extracting audio features. Consequently, we decided to replace it with an approach that produces higher-level audio features to determine if we could achieve similar results. We substituted Wav2vec 2.0 with DeepSpeech \cite{hannun2014deep} feature extraction in the facial animator module and retrained the framework. Using a $29 \times 1$ output with the probabilities of each alphabet as a feature led to a significant increase in the error rate metric and a decrease in lip readability can be observed in Table \ref{table4}. This finding highlights the effectiveness of Wav2vec 2.0 in producing accurate and coherent results.
%We found that Wav2vec 2.0 has excellent capabilities in extracting audio features. Therefore, we replaced it with a lighter-weight approach to determine if it could achieve comparable results. We replaced wav2vec 2.0 with MFCC feature extraction in the facial animator module and retrained the framework. Using MFCC resulted in a significant increase in the error rate metric and a decrease in lip readability. This outcome suggests that MFCC is inadequate for tasks requiring precision and underscores the efficacy of wav2vec 2.0.

%The 29x1 output with the probabilities of each alphabet can be used as a feature in several other tasks.

%We discovered that Wav2vec 2.0 excels in extracting audio features. Consequently, we decided to replace it with a  approach that produce shorter dimension audio features to determine if we could achieve similar results. We substituted Wav2vec 2.0 with DeepSpeech feature extraction in the facial animator module and retrained the framework. Using 29x1  output with the probabilities of each alphabet as a feature led to a significant increase in the error rate metric and a decrease in lip readability. This finding  highlighting the effectiveness of Wav2vec 2.0.

\textbf{Impact of the dual training diagram:} The dual training diagram constitutes the central component of our framework. When we remove the dual training diagram and only train the facial animator module, we observe a significant increase in the error rate of the generated facial animation lip shape, as shown in Table \ref{table4}. 
This observation indicates that our method effectively utilizes limited data to learn more versatile and general representations. The inclusion of the dual training diagram in the framework contributes to improved model performance, resulting in more accurate and expressive facial animations, as evidenced by the lower error rates when the dual training diagram is employed.

\textbf{Impact of the cross-modality consistency regularization loss:} We conducted a comprehensive study to assess the impact of removing the audio and motion consistency regularization loss by eliminating them from our entire model. We calculated the lip distance by finding the midpoints of the upper and lower lips and computing their Euclidean distance. As illustrated in Fig. \ref{fig5}, the cross-modality consistency regularization loss ensures that the lip movements generated by our network are aligned with the original audio. Ignoring these losses resulted in a decrease in the comprehensibility of lip movements. Moreover, the presence of non-verbal cues, which accompany speech and contribute subtle variations to facial animations, is more accurately captured by our model, resulting in the reproduction of fine details and nuances in facial movements, as shown in Table \ref{table4}.
The cross-modality consistency regularization loss enables the model to learn subtle motion and mitigate the over-smoothing issue. The omission of these losses led to a decrease in the visual quality of the generated faces, highlighting the importance of incorporating the audio and motion consistency regularization loss in our framework.

\textbf{Impact of the parameter sharing for face-animation component and lip-reading component encoding and decoding:} Considering the duality of these two components, the encoder and decoder of Audio/Motion can be viewed as reverse transformations of each other. Therefore, we can leverage these properties to propose a corresponding weight-sharing scheme. For input audio embedding, ${A}_{1:T'}$, in the face animation component, the input audio embedding is embedded into features ${X}_{1:T'}$ using the MLP-based audio encoder. For an audio generation in the lip reading component, the predicted audio embedding $\breve{A}_{1:T}$ is decoded to obtain the output through the MLP-based audio decoder. Thus, we can directly share the weights of the encoder and decoder, where the weights of the encoder are required to be the transpose of the decoder. For the input motions in the lip reading component, an MLP is applied to encode the motions into a fixed-size feature vector ${Y}_{1:T}$. For motion generation in the face animation component, an MLP is also applied to decode the vector back to a motion sequence step-by-step. We can also share the parameters of the two MLPs.
However, experimental results show that sharing weights of encodings and decodings will deteriorate the final outcome, indicating that this approach may not be optimal for achieving accurate and coherent results. We assume this may be caused by the larger parameters of the encoder/decoder, which could lead to suboptimal performance when using a weight-sharing scheme.

%Considering the duality of A and B, the encoder and decoder of Audio/Motion can be viewed as reverse transformations of each other. Hence, we could employ these properties to propose a corresponding weight-sharing scheme. For input audio in the A component, the input audio is embedded into features a by the matrix Ea. For the audio generation in the B component, the predicted feature aˆ is decoded for obtaining the answer through a linear classifier Wa. Thus, we can directly share the weights of Ea and Wa, where Ea is required to be the transposition of Wa. For input motions in the B component, MLP is applied to encode the motions into a fixed-size feature vector q. For the motion generation in the A component, MLP is also applied to decode the vector back to a motion sequence step-by-step. We can also share the parameters of the two MLPs. 

%While experimental results show that sharing weights of encoding and decodings will deteriorate the final result. % but requires fewer parameters.

\section{Conclusion}\label{con}

In this paper, we introduced a cross-modal dual learning framework for speech-driven 3D facial animation, effectively tackling the challenges inherent in audio-driven facial animation. By formulating facial animation and lip reading components as dual tasks and incorporating an innovative parameter-sharing scheme and duality regularizer, our method efficiently utilizes data, improves model performance, and generates more accurate and expressive facial animations. Furthermore, the cross-modality consistency regularization loss is crucial for capturing subtle motion and mitigating the over-smoothing issue. 
Experimental results and user studies showed that our methods have a superiority in achieving accurate lip sync and vivid facial expressions and outperform current methods regarding state-of-the-art performance in 3D talking faces, paving the way for future research and advancements in the field of audio-driven facial animation.

%\section*{Acknowledgments}
%This should be a simple paragraph before the References to thank those individuals and institutions who have supported your work on this article.


%\bibliographystyle{plain} 
\bibliographystyle{IEEEtran}
\bibliography{reference}

\section{Appendix}
This supplemental document contains two sections:
Section A shows the implementation details of our DualTalker;
Section B presents details of the user study; 
%and Section D presents short descriptions of the supplemental
%video. T more discussions on the proposed
%method; Section C presents
The source code and trained model will also be released upon publication.

\subsection{Implementation Details}

%A. Implementation Details Network Architecture
\textbf{Network Architecture}
The overall architecture of DualTalker is illustrated in Fig. 1 of the main paper. In this section, we provide more specific parameters adopted for DualTalker trained on the two datasets. As the architectures of the primal and dual tasks are closely related, we will only discuss the facial animation aspect.

For the BIWI dataset, the audio feature extractor in the facial animation task consists of a TCN followed by a linear interpolation layer. This layer down/up samples the input according to the frequency of the captured facial motion data (25fps or 30fps). The interpolated outputs are then fed into 12 identical transformer encoder layers. Each transformer encoder layer has a model dimensionality of 768 and 12 attention heads. The interpolated extracted audio feature is subsequently input into a linear projection-based audio encoder, which converts the 768-dimensional features into d-dimensional speech representations ($d= 256$ for BIWI and $d= 128$ for VOCASET).

The motion encoder is a fully connected layer with $d$ outputs, while the style embedding layer is an embedding layer with $d$ outputs. The cross-modal decoder consists of a motion-attentive (audio-attentive) module, a speaker-aware module, and an audio-visual integration module. For the self-attention-based motion-attentive (audio-attentive) Module, we employ 4 heads and set the model dimensionality to $d$. The style embedding and audio features are then combined along the channel dimension, and this combined data is fed into the speaker-aware module. This module generates Sigmoid-normalized fusion weights for the weighted audio feature. The speaker-aware module comprises a fully connected (FC) layer (converting $d$ to $\frac{2\times{d}}{16}$), a ReLU function, another FC layer, and a Sigmoid function. The cross-modal MH attention-based audio-visual integration module integrates the outputs from the audio encoder (speech features) and the speaker-aware module (speaker-specific motion features). We also employ 4 heads and set the model dimensionality to d for this module. The dimension of the Feed-Forward (FF) layer is 2048. Similar to the encoder, residual connections and layer normalizations are applied to the two biased attention layers and the FF layer. Lastly, a fully connected layer with $v$ outputs serves as the motion decoder ($v = 70110$ for BIWI and $v = 15069$ for VOCASET).

\textbf{Hyper-parameters Setting}
We employ the Adam optimizer with a learning rate of 1e-4. The parameters of the audio feature extractor are initialized using the pre-trained wav2vec 2.0 weights. During training, only the parameters of the TCN are fixed. The parameters of the audio encoder, motion encoder, and style encoder are shared. For the BIWI dataset, we set $\lambda_1 = 1$, $\lambda_2 = 1e-8$, $\lambda_3 = 1e-9$ and $\lambda_4 = 1e-6$, while for the VOCA dataset, we set $\lambda_1 = 1$, $\lambda_2 = 1e-8$ and  $\lambda_3 = 1e-6$.

The models are trained for 100 epochs for BIWI and 100 epochs for VOCA. During the inference process, we only use the primal component to predict facial motion.
\subsection{User Study}
%The designed user interface is shown in Fig.5 . To avoid participants selecting
%an option randomly, we filter out those comparison results completed in less than two and a half minutes. In total, 180 A vs. B pairs (30 samples 6 comparisons) are created for BIWITest-B, and another 180 A vs. B pairs from the VOCA-Test set are created for VOCA-Test. We designed 15 types of questionnaires and the participants will be given the different questionnaires to make sure Each video pair of is judged by at least 3 different participants separately. questionnaire shows sixty video pairs including the qualification test in randomized order and the participants is instructed to judge the videos w.r.t two questions:
%“Comparing the two full faces, which one looks more realistic?” and “Comparing the lips of two faces, which one ismore in sync with audio?”.

\begin{figure*}
  \centering
  \includegraphics[width=0.7\textwidth]{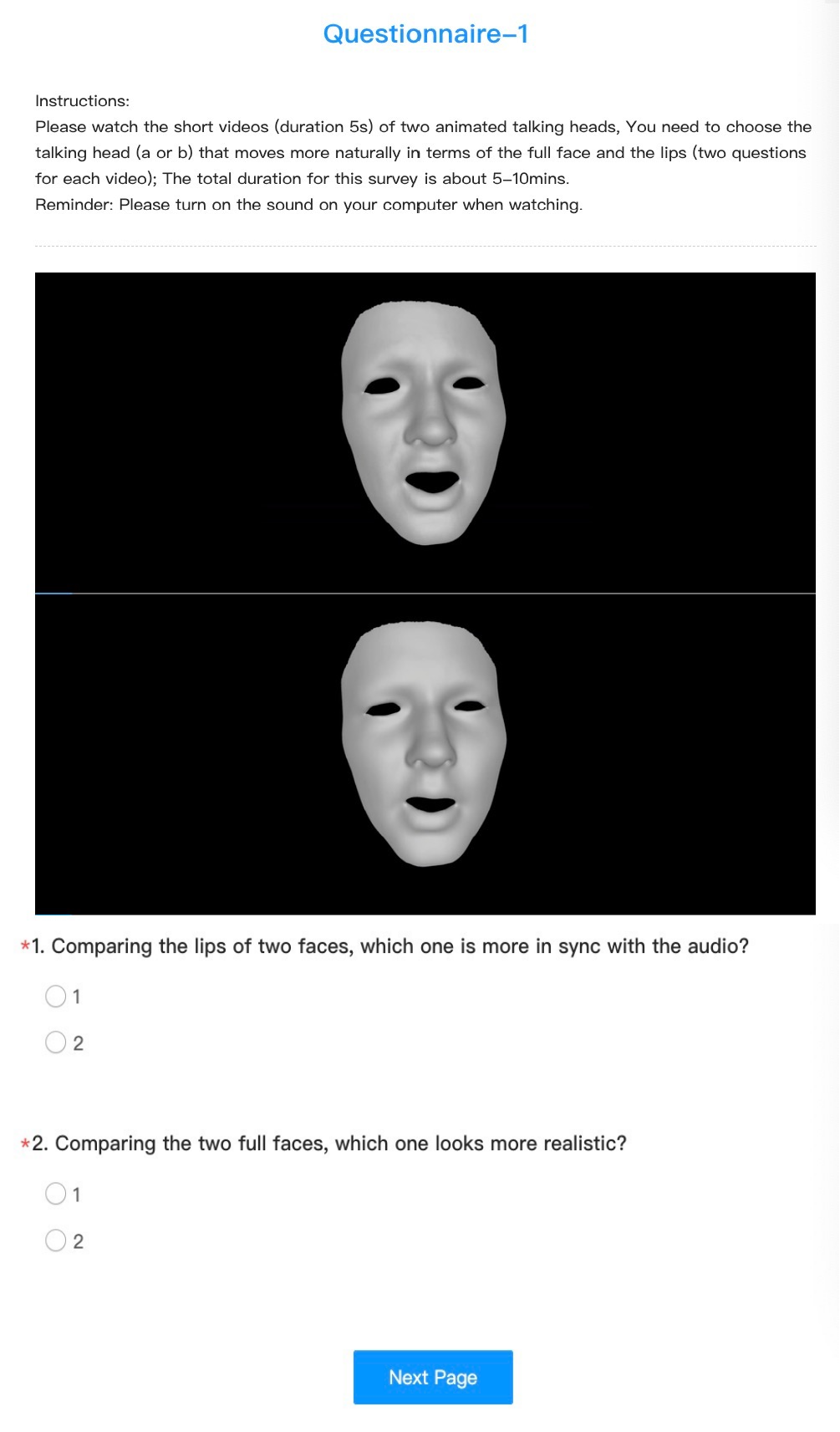}
  \caption{ Example of a designed questionnaire 1 for evaluating the performance of our cross-modal dual learning framework for speech-driven 3D facial animation.}
  \label{fig6}
\end{figure*}

The user interface design is depicted in Fig. \ref{fig6}. To prevent participants from randomly selecting options, we filter out comparison results completed in less than two and a half minutes. In total, 180 A vs. B pairs (30 samples x 6 comparisons) are created for the BIWI-Test-B dataset, and another 180 A vs. B pairs are created for the VOCA-Test dataset. We designed 15 types of questionnaires, and participants are given different questionnaires to ensure each video pair was judged by at least three separate participants. Each questionnaire displays 24 video pairs, including the qualification test, in a randomized order. Participants are instructed to evaluate the videos with respect to two questions: "Comparing the two full faces, which one looks more realistic?" and "Comparing the lips of the two faces, which one is more in sync with the audio?".

\end{document}